%% file: arXiv.tex
\documentclass[runningheads]{llncs}
\usepackage{graphicx}
\usepackage{comment}
\usepackage{amsmath,amssymb} %
\usepackage{color}
\usepackage{enumitem}
\usepackage{booktabs}
\usepackage{bbding}
\usepackage{subcaption}
\usepackage{multirow}
\usepackage{makecell}

\definecolor{citecolor}{RGB}{34, 139, 34}
\usepackage[pagebackref=true,breaklinks=true,colorlinks,bookmarks=false, citecolor=citecolor]{hyperref}

\newcommand{\heading}[1]{
\vspace{1mm}\noindent\textbf{#1}
}

\begin{document}
\pagestyle{headings}
\mainmatter
\def\ECCVSubNumber{3009}

\title{FeatMatch: Feature-Based Augmentation \\ for Semi-Supervised Learning}

\titlerunning{FeatMatch: Feature-Based Augmentation for Semi-Supervised Learning}
\input{sections/authors.tex}
\maketitle

\input{sections/abstract.tex}
\input{sections/introduction.tex}

\input{sections/related_works.tex}

\input{sections/method.tex}

\input{sections/experiments.tex}

\input{sections/conclusion.tex}

\clearpage

\input{sections/appendix}

\bibliographystyle{splncs04}
\bibliography{sections/citations.bib}

\end{document}

%% file: sections/authors.tex
\author{Chia-Wen Kuo\textsuperscript{$\dagger$}, Chih-Yao Ma\textsuperscript{$\dagger$}, Jia-Bin Huang\textsuperscript{$\ddagger$}, Zsolt Kira\textsuperscript{$\dagger$}, \\
\normalsize
\textsuperscript{$\dagger$}Georgia Tech,
\textsuperscript{$\ddagger$}Virginia Tech\\
\vskip 0.1in
\scriptsize
\texttt{albert.cwkuo@gatech.edu},
\texttt{cyma@gatech.edu},
\texttt{jbhuang@vt.edu},
\texttt{zkira@gatech.edu}
}

\authorrunning{Kuo et al.}

%% file: sections/abstract.tex
\begin{abstract}
Recent state-of-the-art semi-supervised learning (SSL) methods use a combination of image-based transformations and consistency regularization as core components. Such methods, however, are limited to simple transformations such as traditional data augmentation or convex combinations of two images. In this paper, we propose a novel learned feature-based refinement and augmentation method that produces a varied set of complex transformations. Importantly, these transformations also use information from both within-class and across-class prototypical representations that we extract through clustering. We use features already computed across iterations by storing them in a memory bank, obviating the need for significant extra computation. These transformations, combined with traditional image-based augmentation, are then used as part of the consistency-based regularization loss. We demonstrate that our method is comparable to current state of art for smaller datasets (CIFAR-10 and SVHN) while being able to scale up to larger datasets such as CIFAR-100 and mini-Imagenet where we achieve significant gains over the state of art (\textit{e.g.,} absolute 17.44\% gain on mini-ImageNet). We further test our method on DomainNet, demonstrating better robustness to out-of-domain unlabeled data, and perform rigorous ablations and analysis to validate the method.

\keywords{semi-supervised learning, feature-based augmentation, consistency regularization}
\end{abstract}

%% file: sections/introduction.tex
\section{Introduction}

\begin{figure}[t]
\begin{subfigure}{\textwidth}
  \centering
  \includegraphics[width=.7\linewidth]{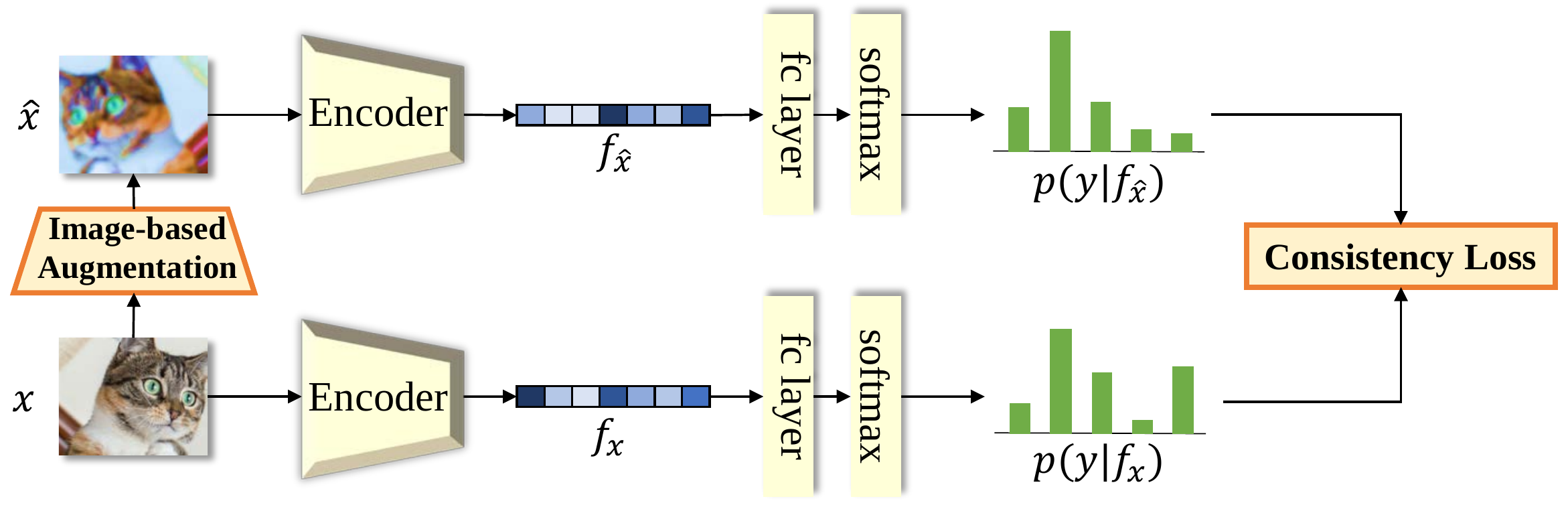}
  \caption{Image-Based Augmentation and Consistency}
  \label{fig:concept-image}
\end{subfigure}

\begin{subfigure}{\textwidth}
  \centering
  \includegraphics[width=.75\linewidth]{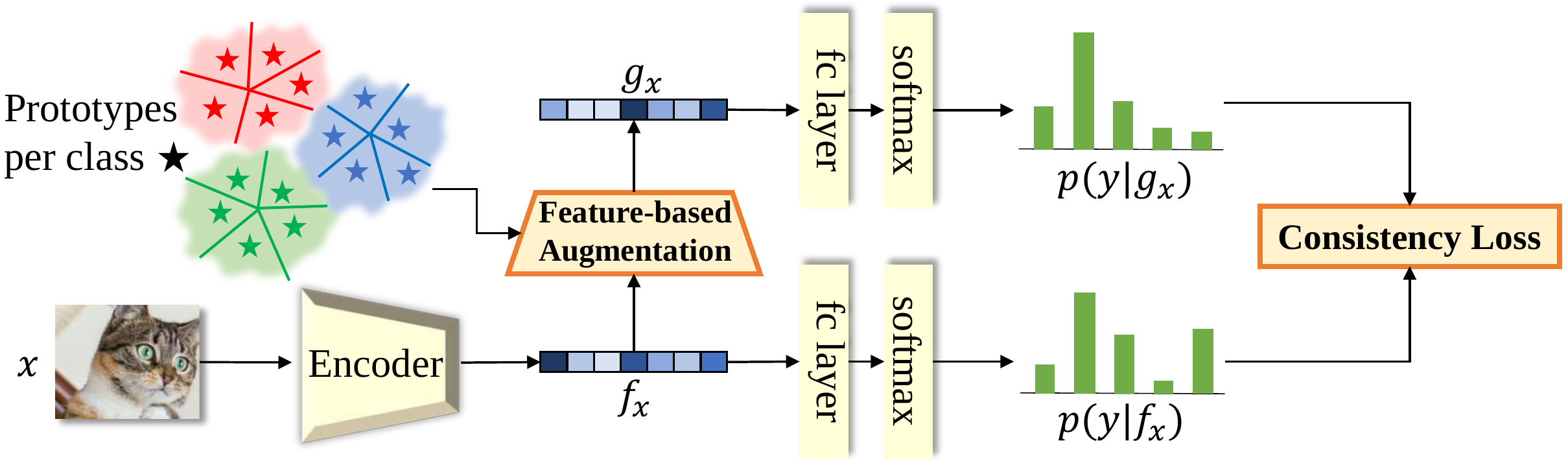}
  \caption{Feature-Based Augmentation and Consistency}
  \label{fig:concept-feature}
\end{subfigure}
\caption{
Consistency regularization methods are the most successful methods for semi-supervised learning.
The main idea of these methods is to enforce consistency between the predictions of different \textit{transformations} of an input image.
(\textit{\textbf{a}}) Image-based augmentation method generate different views of an input image via data augmentation, which are limited to operations in the image space as well as operations within a single instance or simple convex combination of two instances.
(\textit{\textbf{b}})
We propose an additional learned feature-based augmentation that operates in the abstract feature space.
The learned feature refinement and augmentation module is capable of leveraging information from other instances, within or outside of the same class.
}
\label{fig:concept}
\end{figure}

Driven by large-scale datasets such as ImageNet as well as computing resources, deep neural networks have achieved strong performance on a wide variety of tasks.
Training these deep neural networks, however, requires millions of labeled examples that are expensive to acquire and annotate.
Consequently, numerous methods have been developed for semi-supervised learning (SSL), where a large number of unlabeled examples are available alongside a smaller set of labeled data.
One branch of the most successful SSL methods~\cite{Laine2017iclr,miyato2018virtual,qiao2018deep,sajjadi2016regularization,tarvainen2017mean,berthelot2019mixmatch,berthelot2019remixmatch} uses image-based augmentation~\cite{zhang2018mixup,cubuk2019randaugment,hendrycks2020augmix,cubuk2018autoaugment} to generate different $transformations$ of an input image, and consistency regularization to enforce invariant representations across these transformations.
While these methods have achieved great success, the data augmentation methods for generating different transformations are limited to transformations in the image space and fail to leverage the knowledge of other instances in the dataset for diverse transformations.

In this paper, we propose novel feature-based refinement and augmentation that addresses the limitations of conventional image-based augmentation described above.
Specifically, we propose a module that learns to refine and augment input image features via soft-attention toward a small set of representative prototypes extracted from the image features of other images in the dataset.
The comparison between image-based augmentation and our proposed feature-based refinement and augmentation is shown in Fig.~\ref{fig:concept}.
Since the proposed module is learned and carried out in the feature space, diverse and abstract transformations of input images can be applied, which we validate in Sec.~\ref{section:analysis}.
Our approach only requires minimum computation via maintaining a memory bank and using k-means clustering to extract prototypes.

We demonstrate that adding our proposed feature-based augmentation along with conventional image-based augmentations, when used for consistency regularization, achieves significant gains.
We test our method on standard SSL datasets such as SVHN and CIFAR-10, and show that our method, despite its simplicity, compares favorably against state-of-art methods in all cases.
Further, through testing on CIFAR-100 and mini-ImageNet, we show that our method is scalable to larger datasets and outperformed the current best methods by significant margins.
For example, we outperformed the closest state of the art by an absolute
\textbf{17\%} on mini-ImageNet with 4k labels.
We also propose another realistic setting on DomainNet~\cite{peng2019moment} to test the robustness of our proposed method under the case where the unlabeled samples are partially coming from shifted domains, in which we improved \textbf{23\%} over supervised baseline and \textbf{12\%} over semi-supervised baseline when 50\% unlabeled samples are all coming from shifted domains.
Finally, we conduct thorough ablations and thorough analysis to highlight that the method does, in fact, perform varied complex transformations in feature space (as evidenced by t-SNE and nearest neighbor image samples).
  To summarize, our key contributions include:
\begin{itemize}[topsep=0pt,itemsep=-1ex,partopsep=1ex,parsep=1ex,labelindent=0.0em,labelsep=0.2cm,leftmargin=*]
    \item We develop a learned feature-based refinement and augmentation module to transform input image features in the abstract feature space by leveraging a small set of representative prototypes of all classes in the dataset.
    \item We propose a memory bank mechanism to efficiently extract prototypes from images of the entire dataset with minimal extra computations.
    \item We demonstrate thorough results across four standard SSL datasets and also propose a 
    realistic setting where the unlabeled data partially come from domains shifted from the target labeled set.
    \item We perform in-depth analysis of the prototype representations extracted and used for each instance, as well as what transformations the proposed feature-based refinement and augmentation module learns.
\end{itemize}

%% file: sections/related_works.tex
\section{Related Works}
\heading{Consistency Regularization Methods.}
Current state-of-the-art SSL methods mostly fall into this category.
The key insight of this branch of methods is that the prediction of a deep model should be consistent across different \textit{semantic-preserving transformations} of the same data.
Consistency regularization methods regularize the model to be invariant to textural or geometric changes of an image.
Specifically, given an input image $x$ and a network composed of a feature encoder $f_x = Enc(x)$ and a classifier $p_x = Clf(f_x)$, we can generate the pseudo-label of the input image by $p_x = Clf(Enc(x))$.
Furthermore, given a data augmentation module $AugD(\cdot)$, we can generate an augmented copy of $x$ by $\hat{x} = AugD(x)$.
A consistency loss $\mathcal{H}$, typically KL-Divergence loss, is then applied on the model predictions of $\hat{x}$ to enforce consistent prediction: $\mathcal{L}_{con} = \mathcal{H}(p,Clf(Enc(\hat{x})))$.

\heading{Image-Based Augmentation.}
The core to consistency-based methods is how to generate diverse but reasonable transformations of the same data.
A straightforward answer is to incorporate data augmentation, which has been widely used in the training of a deep model to increase data diversity and prevent overfitting.
For example, \cite{berthelot2019mixmatch,Laine2017iclr,sajjadi2016regularization,tarvainen2017mean} use traditional data augmentation to generate different transformations of semantically identical images.
Data augmentation method randomly perturbs an image in terms of its texture, eg. brightness, hue, sharpness, or its geometry, eg. rotation, translation, or affine transform.
In addition to data augmentation, Miyato et al. \cite{miyato2018virtual} and Yu et al. \cite{yu2019tangent} perturbed images along the adversarial direction, and Qiao et al. \cite{qiao2018deep} use multiple networks to generate different views (predictions) of the same data.
Recently, several works propose data augmentation modules for supervised learning or semi-supervised learning, where the augmentation parameters can either be easily tuned \cite{cubuk2019randaugment}, found by RL-training \cite{cubuk2019autoaugment}, or decided by the confidence of network prediction \cite{berthelot2019remixmatch}.

Mixup \cite{zhang2018mixup,yun2019cutmix,yun2019cutmix,hendrycks2020augmix}, similar to data augmentation, is another effective way of increasing data diversity.
It generates new training samples by a convex combination of two images and their corresponding labels.
It has been shown that models trained with Mixup is robust toward out-of-distribution data \cite{guo2019mixup} and is beneficial for the uncertainty calibration of a network \cite{thulasidasan2019mixup}.
Given two images $x_1$ and $x_2$ and their labels (or pseudo labels)  $y_1$ and $y_2$, they are mixed by a randomly sampled ratio $r$ by $\hat{x} = r \cdot x_1 + (1-r) \cdot x_2$ and $\hat{y} = r \cdot y_1 + (1-r) \cdot y_2$. This has been done in feature space as well~\cite{verma2018manifold}.
A standard classification loss $\mathcal{H}(\cdot)$ is then applied on the prediction of the mixed sample $\hat{x}$ and the mixed label $\hat{y}$ by $\mathcal{L}_{mix} = \mathcal{H}(\hat{y}, Clf(Enc((\hat{x})))$.
Originally, Mixup methods were developed for supervised learning.
ICT~\cite{verma2019interpolation} and MixMatch~\cite{berthelot2019mixmatch} introduce Mixup into semi-supervised learning by using the pseudo-label of the unlabeled data.
Furthermore, by controlling the mixing ratio $r$ to be greater than 0.5 as proposed by \cite{berthelot2019mixmatch}, we can make sure that the mixed sample is closer to $x_1$.
Therefore, we can separate the mixed data into labeled mixed batch $\hat{\mathcal{X}}$ if $x_1$ is labeled, and unlabeled mixed batch $\hat{\mathcal{U}}$ if $x_1$ is unlabeled.
Different loss weights can then be applied to modulate the strength of regularization from the unlabeled data.

%% file: sections/method.tex
\input{tables/comparison.tex}

\section{Feature-Based Augmentation and Consistency} \label{section:method}
Image-based augmentation has been shown to be an effective approach to generate different views of an image for consistency-based SSL methods.
However, conventional image-based augmentation has the following two limitations: (1) Operate in image space, which limits the possible transformations to textural or geometric within images, and (2) Operate within a single instance, which fails to transform data with the knowledge of other instances, either within or outside of the same class.
Some recent works that utilize Mixup only partially address the second limitation of conventional data augmentation since mixup operates only between two instances.
On the other hand, Manifold Mixup~\cite{verma2018manifold} approaches the first limitation by performing Mixup in the feature space but is limited to a simple convex combination of two samples.

We instead propose to address these two limitations simultaneously.
We proposed a novel method that refines and augments image features in the abstract feature space rather than image space.
To efficiently leverage the knowledge of other classes, we condense the information of each class into a small set of prototypes by performing clustering in the feature space.
The image features are then refined and augmented through information propagated from prototypes of all classes.
We hypothesize that this feature refinement/augmentation can further improve the feature representations, and these refined features can produce better pseudo-labels than features without the refinement (See Sec.~\ref{section:analysis} for our analysis on this hypothesis).
The feature refinement and augmentation are learned via a lightweight attention network for the representative prototypes and optimized end-to-end with other objectives such as classification loss.
A consistency loss can naturally be applied between the prediction from the original features and the refined features to regularize the network as shown in Fig.~\ref{fig:concept-feature}.

The final model seamlessly combines our novel feature-based augmentation with conventional image-based augmentation for consistency regularization, which is applied to data augmented from both sources.
Despite the simplicity of the method, we find this achieves significant performance improvement.
In summary, we compare our method with other highly relevant SSL works in Table.~\ref{table:method-comparison}.

\subsection{Prototype Selection}

\begin{figure}[t]
\centering
\includegraphics[width=0.85\linewidth]{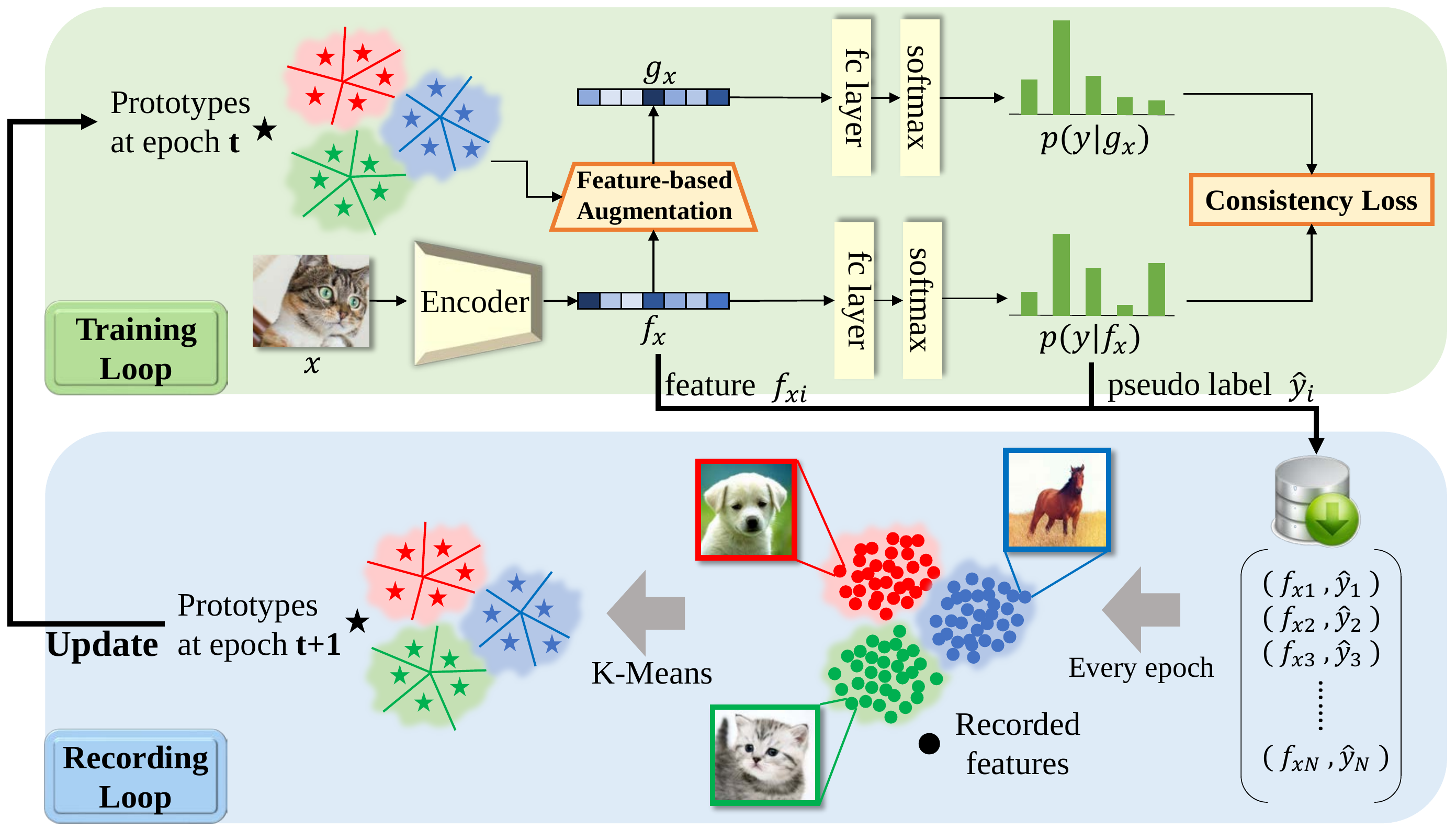}
\caption{
A prototype recording loop that runs alongside the model training loop.
The image features $f_{xi}$ as well as their pseudo labels $\hat{y}_i$ already generated at each iteration of the training loop are collected and recorded in a memory bank as $(f_{xi}, \hat{y}_i)$ pairs.
Once the training loop goes over the whole dataset, the recording loop will run K-Means to extract prototypes for each class, update the prototypes for feature-based augmentation, and clear the memory bank.
}
\label{fig:prototype-selection}
\end{figure}

In order to efficiently leverage the knowledge of other classes for feature refinement and augmentation, we propose to compactly represent the information of each class by clustering in the feature space.
To select representative prototypes from the dataset, we propose to use K-Means clustering in the feature space to extract $p_k$ cluster means as prototypes for \textit{each class}.
However, there are two technical challenges:
(1) in an SSL setting, most images are unlabeled;
(2) even if all the labels are available, it is still computationally expensive to extract features of all the images from the entire dataset before running K-Means.

To tackle these issues, as shown in Fig.~\ref{fig:prototype-selection}, we collect features $f_{xi}$ and pseudo-labels $\hat{y}_i$ already generated by the network at every iteration of the training loop, \textit{i.e.,} no extra computation needed.
In the recording loop, the pairs of pseudo label and features are detached from the computation graph and pushed into a memory bank for later usage.
The prototypes are extracted by K-Means at every epoch when we go over the whole dataset.
Finally, the feature refinement and augmentation module updates the prototypes with the newly extracted ones in the training loop.
Even though the prototypes are extracted from the feature computed from the model a few iterations ago, as training progresses and the model gradually converges, the extracted prototypes fall on the correct cluster and are diverse enough to compactly represent the feature distribution per class.
More analyses can be found in Sec.~\ref{section:analysis}.
Similar idea is concurrently explored in self-supervised learning by He et al.~\cite{wu2018unsupervised,he2019momentum}.

\subsection{Learned Feature Augmentation}

\begin{figure}
\centering
\includegraphics[width=0.85\linewidth]{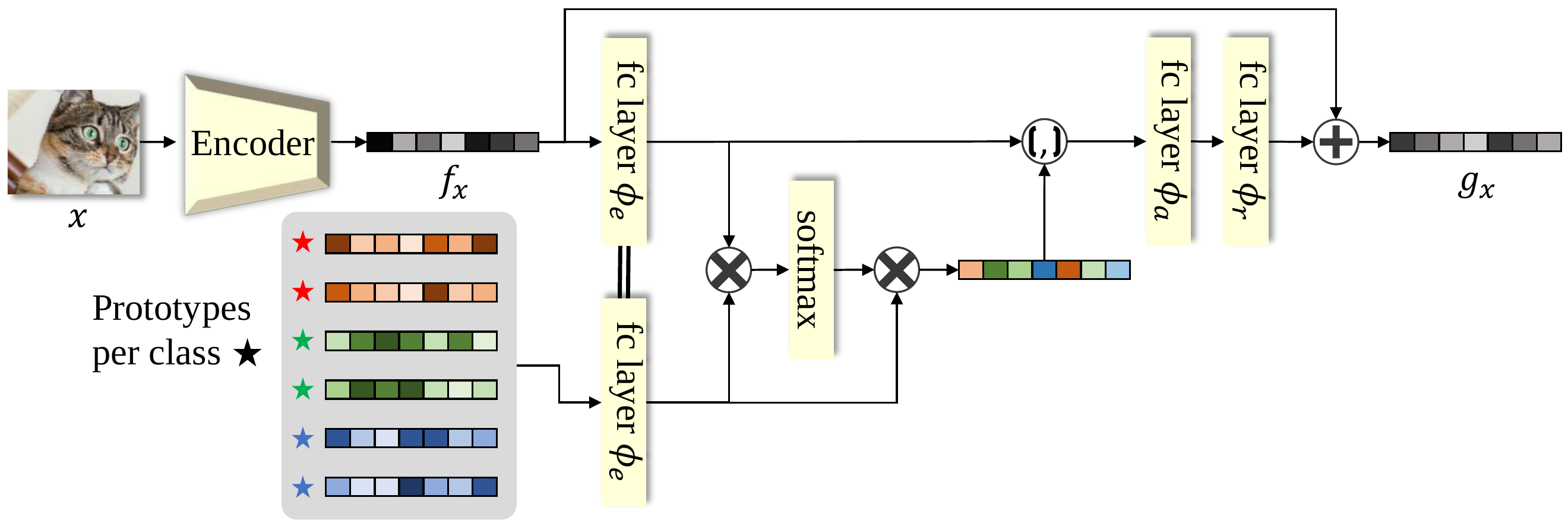}
\caption{
\textbf{Feature-Based Augmentation:} The input image features are augmented by attention using extracted prototype features (Eq.~\ref{eq:attention}), where the colors of $\star$ represent the classes of prototypes.
The prototype features are calcluated via a weighted sum using the attention weights, concatenated with the image features, and then undergo a \textit{fc} layer $\phi_a$ (Eq.~\ref{eq:weighted-sum}) to produce attention features $f_a$.
Finally, we use the attention features to refine and augment the input image features with a residual connection (Eq.~\ref{eq:refine}).
}
\label{fig:feature-augmentation}
\end{figure}

With a set of prototypes selected by the process described above, we propose a learned feature refinement and augmentation module via soft-attention~\cite{vaswani2017attention} toward the set of selected prototypes.
The proposed module refines and augments input image features in the feature space by leveraging the knowledge of prototypes, either within or outside of the same class, as shown in Fig.~\ref{fig:feature-augmentation}.
The lightweight feature refinement
and augmentation module composed of three fully connected layers is jointly optimized with other objectives and hence learns a reasonable feature-based augmentation to aid classification.
We provide further analysis and discussion in Sec.~\ref{section:analysis}.

Inspired by the attention mechanism~\cite{vaswani2017attention}, each input image feature \emph{attends} to prototype features via attention weights computed by dot product similarity.
The prototype features are then weighted summed by the attention weights and then fed back to the input image feature via residual connect for feature augmentation and refinement.
Specifically, for an input image with extracted features $f_x$ and the $i$-th prototype features $f_{p,i}$, we first project them into an embedding space by a learned function $\phi_e$ as $e_x = \phi_e(f_x)$ and $e_{p,i} = \phi_e(f_{p,i})$ respectively.
We compute an attention weight $w_i$ between $e_x$ and $e_{p,i}$ as:
\begin{equation} \label{eq:attention}
    w_i = \text{softmax}(e_x^T e_{p,i}),
\end{equation}
where $\text{softmax}(\cdot)$ normalizes the dot product similarity scores across all prototypes.
The information aggregated from the prototypes and passed to the image features for feature refinement and augmentation can then be expressed as a sum of prototype features weighted by the attention weights:
\begin{equation} \label{eq:weighted-sum}
    f_a = \text{relu}(\phi_a([e_x,\sum_{i} w_i e_{p,i}])),
\end{equation}
where $\phi_a$ is a learnable function, and $[\cdot,\cdot]$ is a concatenation operation along the feature dimension.
Finally, the input image features $f_x$ is refined via a residual connection as:
\begin{equation} \label{eq:refine}
    g_x = \text{relu}(f_x + \phi_r(f_a)),
\end{equation}
where $g_x$ are the refined features of $f_x$, and $\phi_r$ is a learnable function.

The attention mechanism described above can be trivially generalized to multi-head attention as in \cite{vaswani2017attention}.
In practice, we use multi-head attention, instead of single head for slightly better results.
For simplicity, we define the feature refinement and augmentation process $AugF(\cdot)$ described above as $g_x = AugF(f_x)$.

\subsection{Consistency Regularization}\label{sec:consistency-regularization}
The learned $AugF$ module along with the selected prototypes provides an effective method for feature-based augmentation, which addresses the limitations of conventional data augmentation methods discussed previously.
With the learned feature-based augmentation, we can naturally apply a consistency loss between the prediction of unaugmented features $f_x$ and augmented features $g_x$.

However, given a classifier $p = Clf(f)$, which prediction should we use as pseudo-label, $p_g = Clf(g_x)$ or $p_f = Clf(f_x)$?
We investigate this problem in Sec.~\ref{section:analysis} and find that $AugF$ is able to \emph{refine} the input features for better representation, thus generating better pseudo-labels.
Therefore, we compute pseudo-label $p_g$ on the refined feature $g_x$ by $p_g = Clf(g_x)$.
The feature-based consistency loss can be computed as: $\mathcal{L}_{con} = \mathcal{H}(p_g, Clf(f_x))$.
We can easily extend $\mathcal{L}_{con}$%
to work seamlessly with traditional augmentation methods, \textit{i.e.,} traditional data augmentation and Mixup.
For simplicity, we will illustrate with only data augmentation, but Mixup can be easily adapted.
Inspired by Berthelot et al.~\cite{berthelot2019remixmatch}, we generate a weakly augmented image $x$ and its strongly augmented copy $\hat{x}$.
The pseudo-label is computed on the weakly augmented image $x$ that undergoes feature-based augmentation and refinement for better pseudo-labels as $p_g = Clf(AugF(Enc(x)))$.
We can then compute two consistency losses on the strongly augmented data $\hat{x}$, one with $AugF$ applied and the other without:
\begin{equation}\label{eq:con-g}
    \mathcal{L}_{con\text{-}g} = \mathcal{H}(p_g, Clf(AugF(Enc(\hat{x})))
\end{equation}
\begin{equation}\label{eq:con-f}
    \mathcal{L}_{con\text{-}f} = \mathcal{H}(p_g, Clf(Enc(\hat{x})))
\end{equation}
Since the pseudo-label $p_g$ is computed on the image undergoing weak data augmentation and feature-based augmentation, the regularization signal of $\mathcal{L}_{con\text{-}g}$ and $\mathcal{L}_{con\text{-}f}$ comes from both image-based and feature-based augmentation.

\subsection{Total Loss}
Consistency regularization losses $\mathcal{L}_{con\text{-}g}$ and $\mathcal{L}_{con\text{-}f}$ in Eq.~\ref{eq:con-g} and \ref{eq:con-f} are applied on unlabeled data.
For labeled image $x$ with label $y$, a regular classification loss can be applied:
\begin{equation}\label{eq:clf}
    \mathcal{L}_{clf} = \mathcal{H}(y, Clf(AugF(Enc(x))))
\end{equation}
Therefore, the total loss can be written as: $\mathcal{L}_{clf} + \lambda_g\mathcal{L}_{con\text{-}g} + \lambda_f\mathcal{L}_{con\text{-}f}$.
Where $\lambda_g$ and $\lambda_f$ are weights for $\mathcal{L}_{con\text{-}g}$ and $\mathcal{L}_{con\text{-}f}$ losses respectively.

%% file: tables/comparison.tex
\begin{table*}[t]
\centering
\renewcommand{\arraystretch}{1.2}
\caption{
Comparison to other SSL methods with consistency regularization.
}
\label{table:method-comparison}
\resizebox{0.85\textwidth}{!}{
\begin{tabular}{@{\extracolsep{4pt}}lcccccc@{}}
\toprule
& \makecell{ReMixMatch\\\cite{berthelot2019remixmatch}} & \makecell{MixMatch\\\cite{berthelot2019mixmatch}} & \makecell{Mean Teacher\\\cite{tarvainen2017mean}} & \makecell{ICT\\\cite{verma2019interpolation}} & \makecell{PLCB\\\cite{arazo2019pseudo}} & \makecell{\textbf{FeatMatch}\\(Ours)}\\
\midrule
\textbf{Feature-Based Augmentation} & -              & -              & -              & -              & -              & \CheckmarkBold \\
Image-Based Augmentation            & \CheckmarkBold & \CheckmarkBold & \CheckmarkBold & \CheckmarkBold & \CheckmarkBold & \CheckmarkBold \\
Temporal Ensembling                 & \CheckmarkBold & \CheckmarkBold & \CheckmarkBold &                &                & -              \\
Self-Supervised Loss                & \CheckmarkBold & -              & -              & -              & -              & -              \\
Alignment of Class Distribution     & \CheckmarkBold & -              & -              & -              & \CheckmarkBold & -              \\
\bottomrule
\end{tabular}
}
\end{table*}

%% file: sections/experiments.tex
\section{Experiments}
\subsection{Datasets}
\subsubsection{Standard SSL datasets.}
We conduct experiments on commonly used SSL datasets: SVHN~\cite{netzer2011reading}, CIFAR-10~\cite{krizhevsky2009learning}, CIFAR-100~\cite{krizhevsky2009learning}, and mini-ImageNet~\cite{ravi2016optimization}.
Following the standard approach in SSL, we randomly choose a certain number of labeled samples as a small labeled set and discard the labels for the remaining data to form a large unlabeled set.
Our proposed method is tested under various amounts of labeled samples.
SVHN is a dataset of 10 digits, which has about 70k training samples.
CIFAR-10 and CIFAR-100 are natural image datasets with 10 and 100 classes respectively.
Both dataset contains 50k training samples.
For mini-ImageNet, we follow \cite{iscen2019label,arazo2019pseudo} to construct the mini-ImageNet training set.
Specifically, given a predefined list of 100 classes~\cite{ravi2016optimization} from ILSVRC~\cite{russakovsky2015imagenet}, 500 samples are selected randomly for each class, thus forming a training set of 50k samples.
The samples are center-cropped and resized to 84x84 resolution.
We then follow the same standard procedure and construct a small labeled set and a large unlabeled set from the 50k training samples.

\input{tables/sota_cifar100_mimagenet.tex}

\noindent
\textbf{SSL under domain shift.}
In another realistic setting, we argue that the unlabeled data may come from a domain different from that of the target labeled data.
For instance, given a small set of labeled natural images of animals, the large unlabeled set may also
contain paintings of animals.
To investigate the effect of domain shift in the unlabeled set, we proposed a new SSL task based on the DomainNet dataset~\cite{peng2019moment}, which contains 345 classes of images coming from six domains: Clipart, Infograph, Painting, Quickdraw, Real, and Sketch.

We use the \emph{Real} domain as our target. Five percent of the data from the Real domain are kept as the target labeled set, and the rest are the target unlabeled set.
We select \emph{Clipart}, \emph{Painting}, \emph{Sketch}, and \emph{Quickdraw} as shifted domains.
To modulate the level of domain shift in the unlabeled data, we propose a parameter $r_u$ that controls the ratio of unlabeled data coming from the target Real domain or the shifted domains.
Specifically, $r_u$ percent of target Real unlabeled set is replaced with data uniformly drawn from the shifted domains.
By formulating the problem this way, the amount of unlabeled data remains constant.
The only factor that affects the performance of the proposed method is the ratio between in-domain data and shifted domain data in the unlabeled set.

We randomly reserve 1\% of data from the Real domain as the validation set.
The final result is reported on the test set of the Real domain, with the model selected on the reserved validation set.
The images are center-cropped and resized to 128x128 resolution, and the model we use is the standard ResNet-18~\cite{he2016deep}.
There are around 120k training samples, which is more than twice larger than the standard SSL datasets such as CIFAR-10 and CIFAR-100.
For a fair comparison, we fix  \emph{all} hyper-parameters across experiments of different $r_u$ to truly assess the robustness of proposed methods toward domain shift in the unlabeled data.

\noindent
\textbf{Hyper-parameters.}
We tune the hyper-parameters on CIFAR-10 with 250 labels with a validation set held-out from the training set.
Our method is not sensitive to the hyper-parameters, which are kept fixed across \emph{all} the datasets and settings.
Please see the supplementary for more implementation details and the values of hyper-parameters.

\input{tables/sota_domainnet.tex}

\subsection{Results}\label{section:results}

We first show our results on CIFAR-100 and mini-ImageNet with 4k and 10k labels in Table~\ref{table:cifar100-mimagenet}.
Our method consistently improves over state of the arts by large margins, with about absolute 5\% on CIFAR-100 with 4k labels and 17\% on mini-ImageNet with 4k labels.

In Table~\ref{table:domainnet}, we show our results on the larger dataset of DomainNet setting, which contains unlabeled data coming from other shifted domains.
It can be clearly seen that in the setting of $r_u = 50\%$, where 50\% of the unlabeled data are coming from other shifted domains, the performance drops by a large margin compared with the setting of $r_u = 0\%$, where all the unlabeled data are coming from the same domain as the target labeled set.
Nevertheless, our proposed feature-based augmentation method improves over supervised baseline by absolute 36\% error rate when $r_u = 0\%$ and 23\% when $r_u = 50\%$.
When compared to the conventional image-based augmentation baseline, we improves by 12\% when $r_u = 50\%$ and 16\% when $r_u = 0\%$.

In Table~\ref{table:cifar10-svhn}, we show the comparison of our method with other SSL methods on standard CIFAR-10 and SVHN datasets.
Our method achieves comparable results with the current state of the art, ReMixMatch, even though 1) we start from a lower baseline and 2) our method is much simpler (\textit{e.g.,} no class distribution alignment and no self-supervised loss), as compared in Table~\ref{table:method-comparison}.
Our proposed feature-based augmentation method is complementary to image-based methods and can be easily integrated to further improve the performance.

\input{tables/sota_cifar10_svhn.tex}
\input{tables/ablation.tex}

\subsection{Ablation Study}

In the ablation study, we are interested in answering the following questions:
1) what is the effectiveness of the two proposed consistency losses -- $\mathcal{L}_{con\text{-}f}$ (Eq.~\ref{eq:con-f}) and $\mathcal{L}_{con\text{-}g}$ (Eq.~\ref{eq:con-g}).
2) how much of the improvement is from the proposed feature-based augmentation method over the image-based augmentation baseline?
For the image-based augmentation baseline, the $AugF$ module is completely removed and thus the consistency regularization comes only from image-based augmentation.
This is also the same image-based augmentation baseline that our final model with feature-based augmentation builds upon.
The ablation study is conducted on CIFAR-10 with various amount of labeled samples (Table~\ref{table:ablation}).

We can see from Table~\ref{table:ablation} that our image-based augmentation baseline achieves good results but only on cases where there are more labeled samples.
We conjecture this is because the aggressive data augmentation applied to training images makes the training unstable.
Nevertheless, our baseline performance is still competitive with respect to other image-based augmentation methods in Table~\ref{table:cifar10-svhn} (though slightly worse than MixMatch).
By adding our proposed $AugF$ module($\mathcal{L}_{con\text{-}f}$ and $\mathcal{L}_{con\text{-}g}$) for feature refinement and augmentation on top of the image-based augmentation baseline, the performance improves over baseline consistently, especially for 250 labels. 

We can also see that $\mathcal{L}_{con\text{-}f}$ plays a more important role than $\mathcal{L}_{con\text{-}g}$, though our final model with both loss terms achieves the best result.
In both $\mathcal{L}_{con\text{-}f}$ and $\mathcal{L}_{con\text{-}g}$, the pseudo-labels are computed from the features undergone feature-based augmentation.
The only difference is which prediction we're driving to match the pseudo-label: 1) the prediction from the feature undergone both $AugD$ and $AugF$ (by $\mathcal{L}_{con\text{-}g}$ loss), or 2) the prediction from the feature undergone only $AugD$ (by $\mathcal{L}_{con\text{-}f}$ loss)?
As claimed in Sec.~\ref{sec:consistency-regularization} and analyzed in Sec.~\ref{section:analysis}, $AugF$ is able to refine input image features for better representation and pseudo-labels of higher quality.
Therefore, matching the slightly worse prediction from the feature undergone only $AugD$ (by $\mathcal{L}_{con\text{-}f}$ loss) induces a stronger consistency regularization.
This explains why $\mathcal{L}_{con\text{-}f}$ improves performance more crucially.

\begin{figure}[t]
\centering
\begin{subfigure}{0.51\textwidth}
  \centering
  \includegraphics[width=\linewidth]{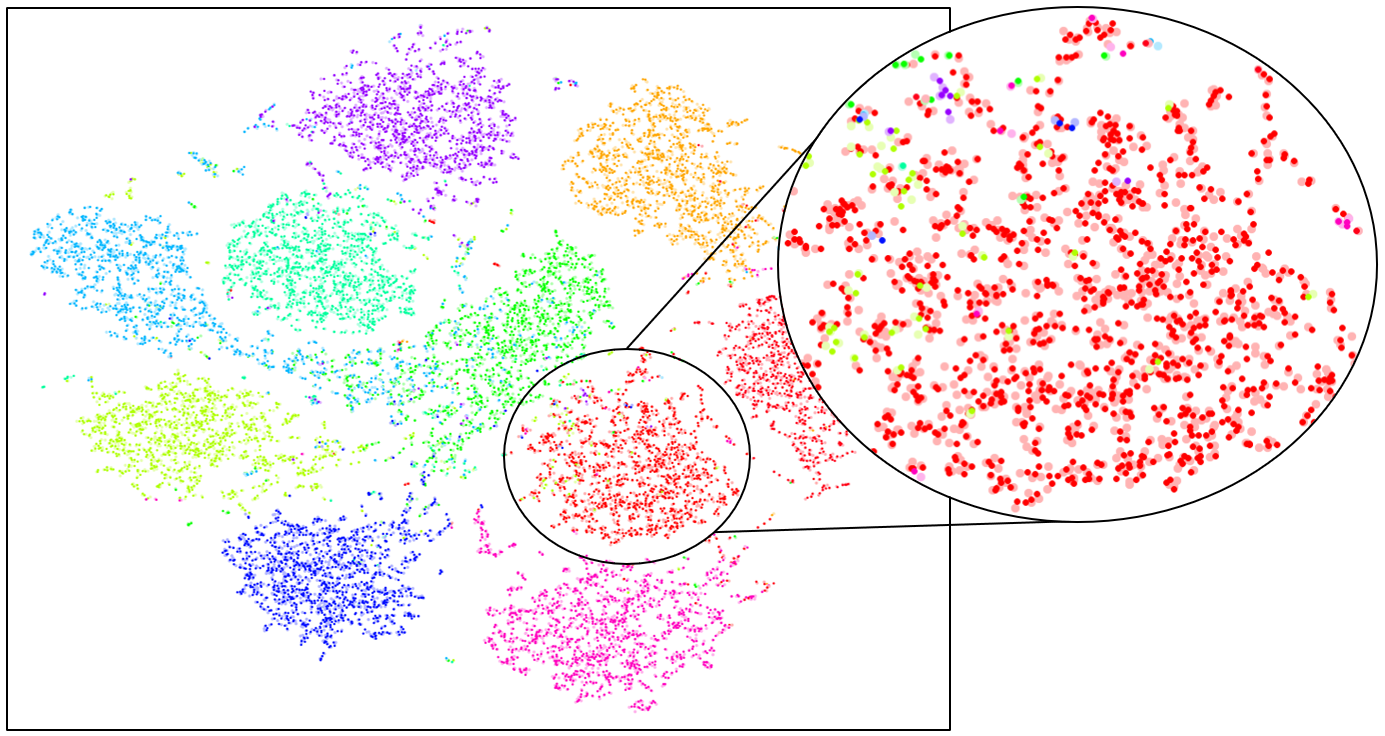}
  \caption{t-SNE of AugD}
  \label{fig:analysis-data-augmentation-tsne}
\end{subfigure}
\begin{subfigure}{0.44\textwidth}
  \centering
  \includegraphics[width=\linewidth]{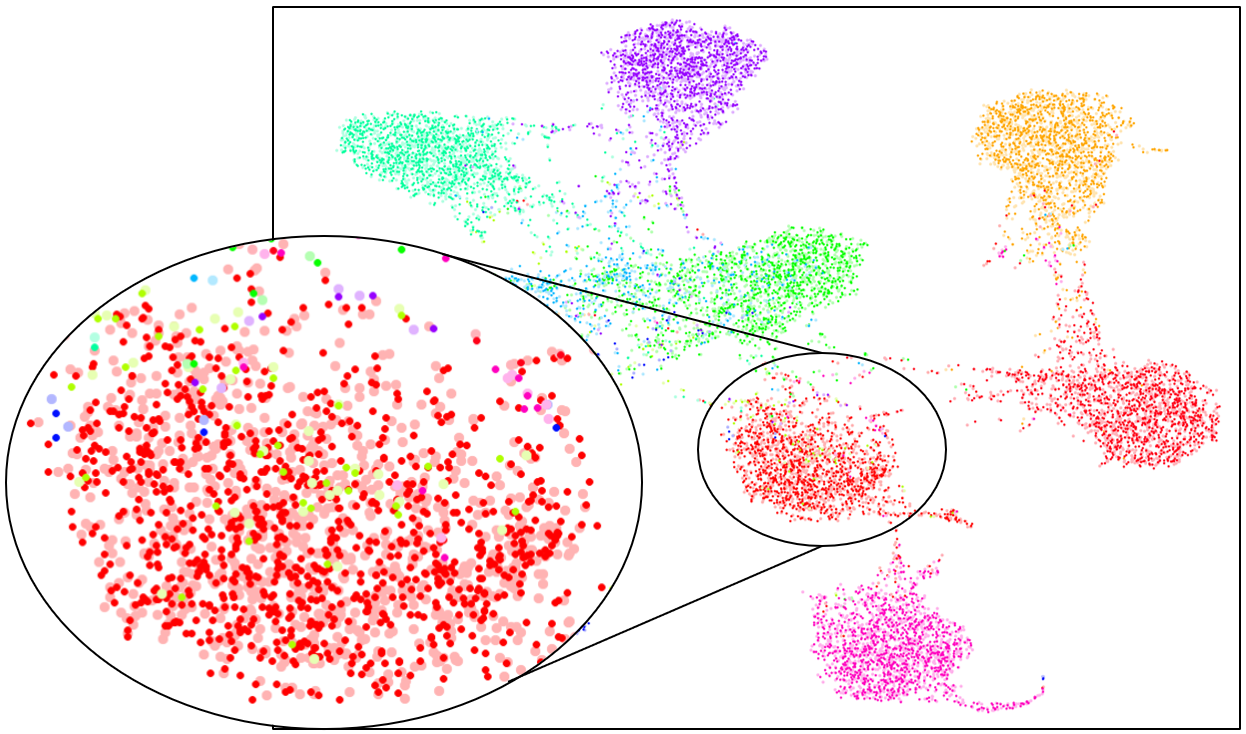}
  \caption{t-SNE of AugF}
  \label{fig:analysis-feature-augmentation-tsne}
\end{subfigure}

\begin{subfigure}{0.9\textwidth}
  \centering
  \includegraphics[width=\linewidth]{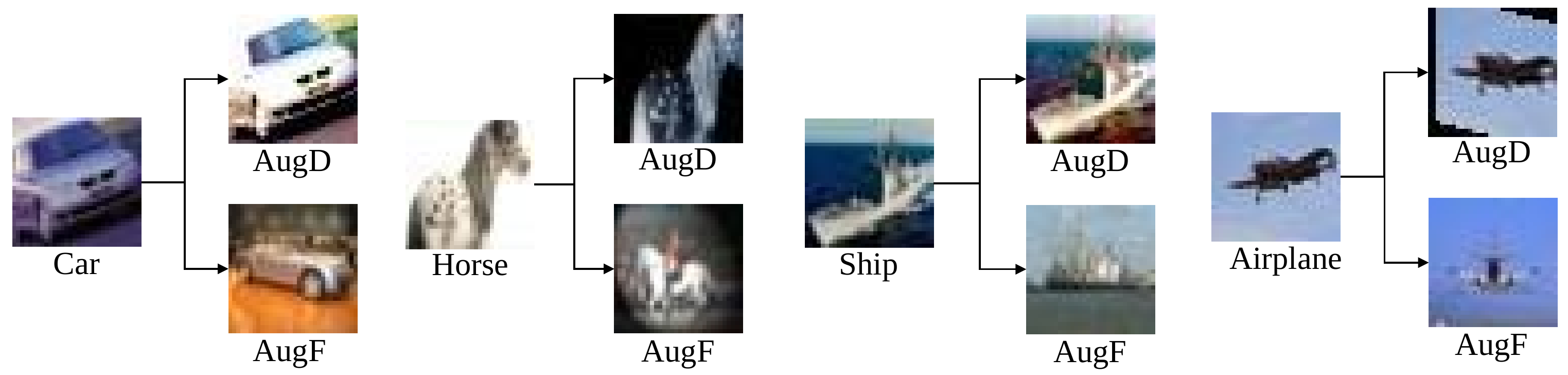}
  \caption{Augmented images of AugD vs. AugF}
  \label{fig:analysis-augmented-image}
\end{subfigure}
\caption{
(\textit{\textbf{a}}) We jointly compute and plot t-SNE of input unaugmented image features (dimmer color) and image-based augmented features (brighter color).
(\textit{\textbf{b}}) We also jointly compute and plot t-SNE of input unaugmented image features (dimmer color) and feature-based augmented features (brighter color) with the exact same t-SNE parameters with (a).
(\textit{\textbf{c}}) To concretely visualize the augmented feature, we find their nearest image neighbor in the feature space and compare against the image-based augmentation method side by side.
}
\label{fig:analysis-augmentation}
\end{figure}

\begin{figure}[t]
\centering
\begin{subfigure}{0.51\textwidth}
  \centering
  \includegraphics[width=\linewidth]{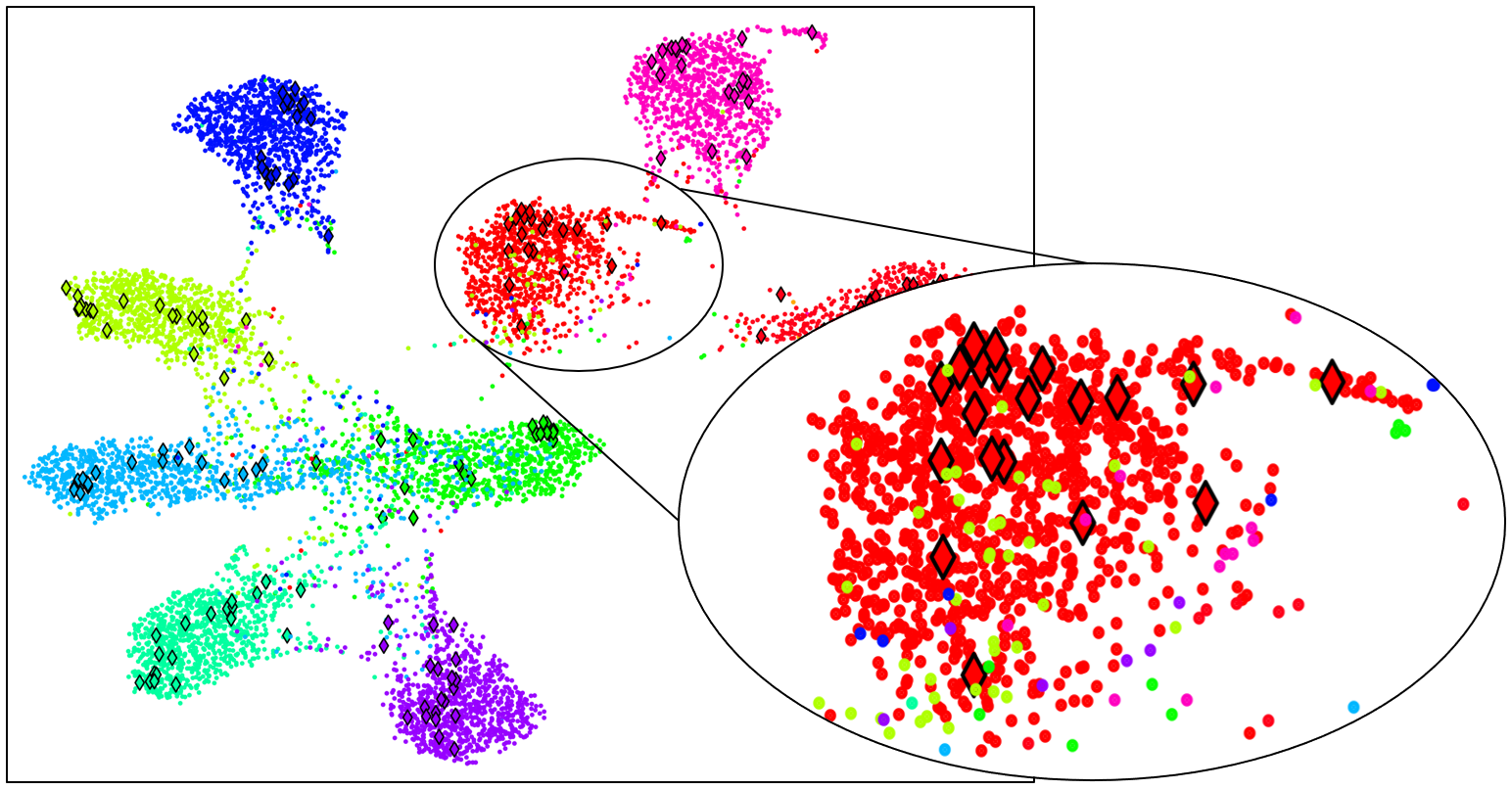}
  \caption{t-SNE of selected prototypes.}
  \label{fig:analysis-prototype-tsne}
\end{subfigure}
\begin{subfigure}{0.36\textwidth}
  \centering
  \includegraphics[width=\linewidth]{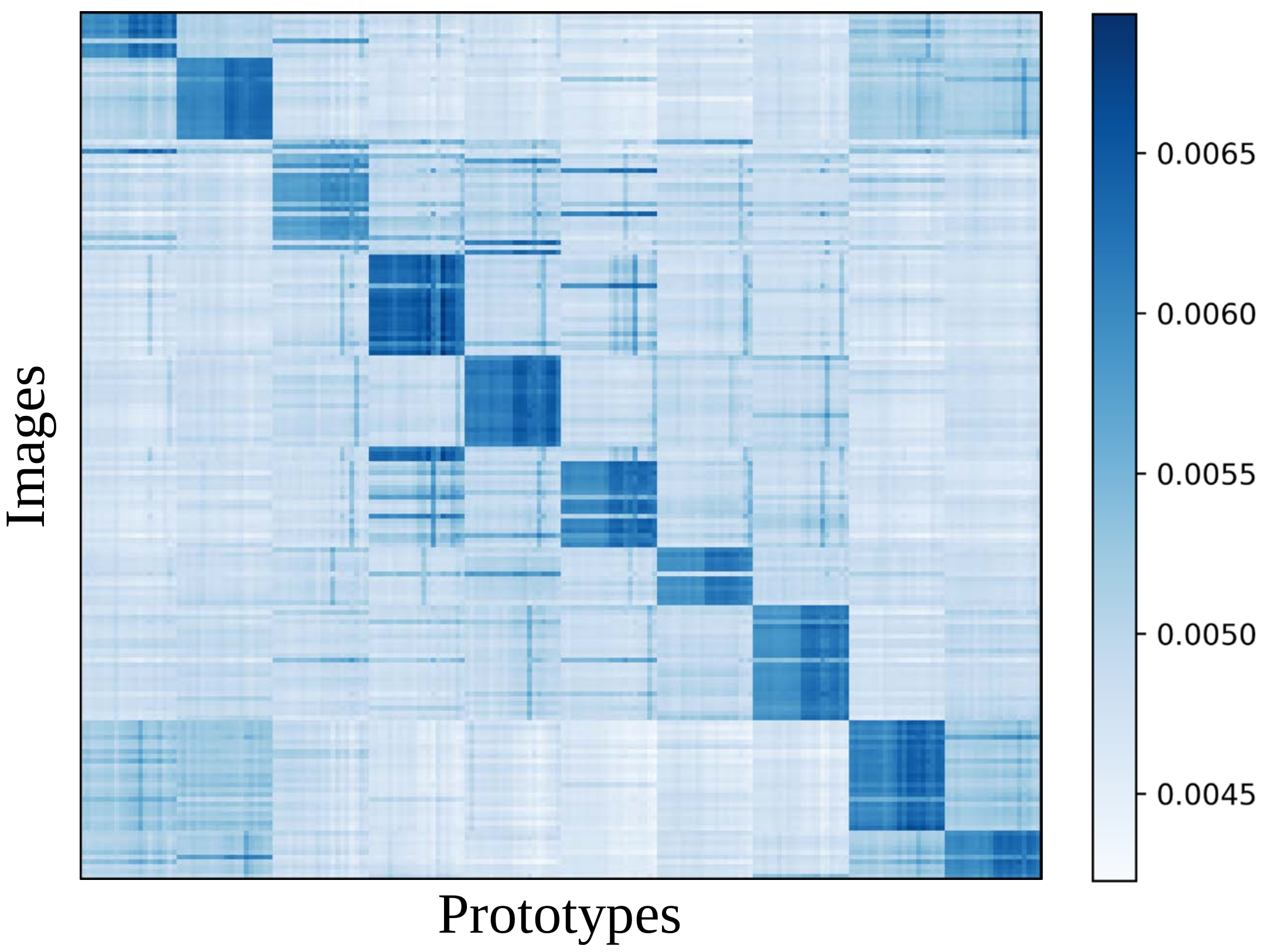}
  \caption{Leaned attention weights.}
  \label{fig:analysis-attention}
\end{subfigure}

\begin{subfigure}{0.9\textwidth}
  \centering
  \includegraphics[width=\linewidth]{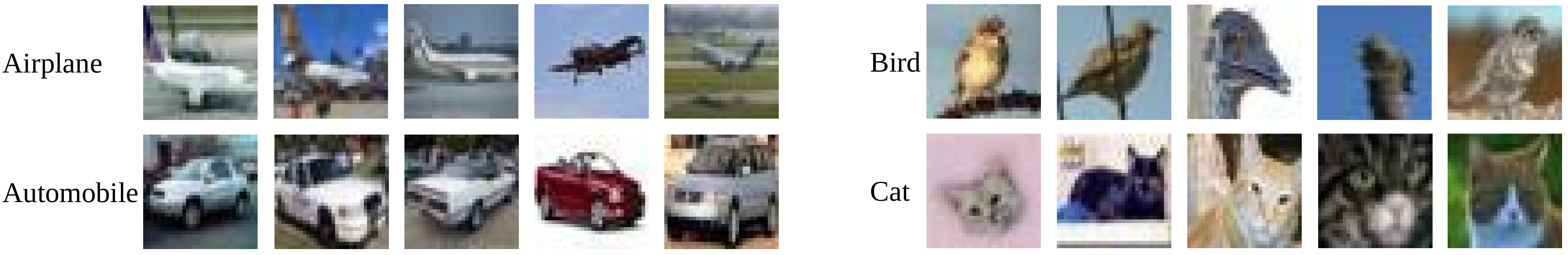}
  \caption{Nearest image neighbors of prototypes}
  \label{fig:analysis-prototype-image}
\end{subfigure}
\caption{
(\textit{\textbf{a}}) In the t-SNE plot, the extracted prototypes ($\diamond$) fall on the correct clusters and are able to compactly represent the cluster.
(\textit{\textbf{b}}) We visualize the learned attention weights from a batch of images toward prototypes.
The images and prototypes are sorted by their classes for ease of illustration.
As can be seen, images have higher attention weights to the prototypes with the same class.
(\textit{\textbf{c}}) We find the prototypes' nearest image neighbors in the feature space.
The prototypes compactly represent a diverse sets of images in each class.
}
\label{fig:analysis-augf}
\end{figure}

\subsection{Analysis}\label{section:analysis}
\textbf{What augmentation does \textit{AugF} learn?}
We compare the feature distribution via t-SNE 1) between input unaugmented image features and image-based augmented features in Fig.~\ref{fig:analysis-data-augmentation-tsne}, and 2) between input unaugmented image features and feature-based augmented features in Fig.~\ref{fig:analysis-feature-augmentation-tsne}.
In Fig.~\ref{fig:analysis-data-augmentation-tsne}, some local small clusters are captured by t-SNE and can be found in the zoomed sub-figure.
This indicates that $AugD$ can only perturb data locally, and fail to produce stronger augmentation for more effective consistency regularization in the feature space.
In Fig.~\ref{fig:analysis-feature-augmentation-tsne}, we can see $AugF$ indeed learns to augment and refine features.
Furthermore, the learned augmentation preserves semantic meaning as the augmented features still fall in the correct cluster.
In the zoomed figure, we can see that the perturbed features distribute more uniformly and no local small clusters could be found.
This indicates that $AugF$ can produce stronger augmentation for more effective consistency regularization in the feature space.

To have a more concrete sense of the learned feature-based augmentation ($AugF)$, we show the augmented feature's nearest image neighbor in the feature space.
Some sample results are shown in Fig.~\ref{fig:analysis-augmented-image}, with the comparison to image-based augmentation ($AugD$) side by side.
As shown in the figure, $AugF$ is capable of transforming features in an abstract way, which goes beyond simple textural and geometric transformation as $AugD$ does.
For instance, it is able to augment data to different poses and backgrounds, which could be challenging for conventional image-based augmentation methods.

\noindent
\textbf{What other reason does \textit{AugF} improve model performance?}
We hypothesize that one other reason why our method can improve performance is that $AugF$ module is capable of refining input image features for better representation by the extracted prototypes, and thus provides better pseudo-labels.
The consistency regularization losses then drive the network's prediction to match the target pseudo-labels of higher quality, leading to overall improvement.
With this hypothesis, we expect classification accuracy to be higher for features after feature refinement.
To verify, we remove $\mathcal{L}_{con\text{-}f}$ loss and retrain.
The accuracy of pseudo-labeling from the features refined by $AugF$ is on average $0.5-1.0 \%$ higher.
This confirms our hypothesis that $\mathcal{L}_{con\text{-}f}$ drives the feature encoder to learn a better feature representation refined by $AugF$.

The reader may wonder: why doesn't $AugF$ learn a shortcut solution of identity mapping to minimize $\mathcal{L}_{con\text{-}f}$ and $\mathcal{L}_{con\text{-}g}$?
As can be seen from Fig.~\ref{fig:analysis-augmentation}, $AugF$ does \emph{not} learn an identity mapping.
Although learning an identity mapping may be a shortcut solution for minimizing $\mathcal{L}_{con\text{-}f}$ and $\mathcal{L}_{con\text{-}g}$, it is not the case for the classification loss $\mathcal{L}_{clf}$ (Eq.~\ref{eq:clf}).
This finding implicitly confirms our hypothesis that there is extra information from the prototypes that $AugF$ can leverage to refine the feature representation for higher (pseudo-label) classification accuracy.

\noindent
\textbf{What does \textit{Aug} do internally?}
In Fig.~\ref{fig:analysis-prototype-tsne} and \ref{fig:analysis-prototype-image}, we can see that even though our proposed prototype extraction method only uses simple K-Means to extract prototypes of each class based on potentially noisy pseudo-labels, and features recorded several iterations ago, our prototype selection method can still successfully extract a diverse set of prototypes per class.
Moreover, in Fig.~\ref{fig:analysis-attention}, the attention mechanism inside $AugF$ learns to attend to prototypes that belong to the same class with the input image feature.
Note that there is no loss term specific for $AugF$, as it is simply jointly optimized with the standard classification and consistency regularization loss from semi-supervised learning.

%% file: tables/sota_cifar100_mimagenet.tex
\begin{table*}[t]
\centering
\renewcommand{\arraystretch}{1.2}
\caption{
Comparison on CIFAR-100 and mini-imageNet. Numbers represent error rate in three runs. For fair comparison, we use the same model as other methods: CNN-13 for CIFAR-100 and ResNet-18 for mini-ImageNet.
}
\label{table:cifar100-mimagenet}
\resizebox{0.7\textwidth}{!}{
\begin{tabular}{@{\extracolsep{4pt}}lcccc@{}}
\toprule
&  \multicolumn{2}{c}{\textbf{CIFAR-100}} & \multicolumn{2}{c}{\textbf{mini-ImageNet}} \\
&  \multicolumn{2}{c}{\# Labeled samples} & \multicolumn{2}{c}{\# Labeled samples} \\
\cline{2-3} \cline{4-5}
Method & 4,000 & 10,000 & 4,000 & 10,000 \\
\midrule
$\Pi$-model~\cite{sajjadi2016regularization}        & -                & 39.19 $\pm$ 0.36 & - & - \\
SNTG~\cite{luo2018smooth}                           & -                & 37.97 $\pm$ 0.29 & - & - \\
SSL with Memory~\cite{chen2018semi}                 & -                & 34.51 $\pm$ 0.61 & - & - \\
Deep Co-Training~\cite{qiao2018deep}                & -                & 34.63 $\pm$ 0.14 & - & - \\
Weight Averaging~\cite{athiwaratkun2018improving}   & -                & 33.62 $\pm$ 0.54 & - & - \\
Mean Teacher~\cite{tarvainen2017mean}               & 45.36 $\pm$ 0.49 & 36.08 $\pm$ 0.51 & 72.51 $\pm$ 0.22 & 57.55 $\pm$ 1.11 \\
Label Propagation~\cite{iscen2019label}             & 43.73 $\pm$ 0.20 & 35.92 $\pm$ 0.47 & 70.29 $\pm$ 0.81 & 57.58 $\pm$ 1.47 \\
PLCB~\cite{arazo2019pseudo}                         & 37.55 $\pm$ 1.09 & 32.15 $\pm$ 0.50 & 56.49 $\pm$ 0.51 & 46.08 $\pm$ 0.11 \\
FeatMatch (Ours)                                    & \textbf{31.06 $\pm$ 0.41} & \textbf{26.83 $\pm$ 0.04} & \textbf{39.05 $\pm$ 0.06} & \textbf{34.79 $\pm$ 0.22} \\
\bottomrule
\end{tabular}
}
\end{table*}

%% file: tables/sota_domainnet.tex
\begin{table*}[t]
\centering
\renewcommand{\arraystretch}{1.2}
\caption{
Comparison between the image-based baseline with our proposed feature-based augmentation method on DomainNet with 1) unlabeled data coming from the same domain as the labeled target ($r_u = 0 \%$), and 2) half of unlabeled data coming from the same domain as the labeled target and the other half from shifted domains ($r_u = 50 \%$).
Numbers are error rates across 3 runs.
}
\label{table:domainnet}
\resizebox{0.72\textwidth}{!}{
\begin{tabular}{@{\extracolsep{4pt}}lcc@{}}
\toprule
Method ($5\%$ labeled samples) & $r_u = 0 \%$ & $r_u = 50 \%$ \\
\midrule
(Semi-supervised) Baseline         & 56.63 $\pm$ 0.17 & 65.82 $\pm$ 0.07 \\
FeatMatch (Ours) & 40.66 $\pm$ 0.60 & 54.01 $\pm$ 0.66 \\ \hline \hline
Supervised baseline ($5\%$ labeled samples, lower bound, ) & \multicolumn{2}{c}{77.25 $\pm$ 0.52} \\
Supervised baseline ($100\%$ labeled samples, upper bound) & \multicolumn{2}{c}{31.91 $\pm$ 0.15} \\
\bottomrule
\end{tabular}
}
\end{table*}

%% file: tables/sota_cifar10_svhn.tex
\begin{table*}[t]
\centering
\renewcommand{\arraystretch}{1.2}
\caption{
Comparison on CIFAR-10 and SVHN. Numbers represent error rate across three runs.
The results reported in the first block with CNN-13 model~\cite{Laine2017iclr,miyato2018virtual} are from the original paper.
The results reported in the second block with wide ResNet (WRN) are reproduced by \cite{berthelot2019mixmatch,berthelot2019remixmatch}.
}
\label{table:cifar10-svhn}
\resizebox{\textwidth}{!}{
\begin{tabular}{@{\extracolsep{4pt}}lccccccc@{}}
\toprule
&  & \multicolumn{3}{c}{\textbf{CIFAR-10}} & \multicolumn{3}{c}{\textbf{SVHN}} \\
&  & \multicolumn{3}{c}{\# Labeled samples} & \multicolumn{3}{c}{\# Labeled samples} \\
\cline{3-5} \cline{6-8}
Method & Model (param.) & 250 & 1,000 & 4,000 & 250 & 1,000 & 4,000\\
\midrule
SSL with Memory~\cite{chen2018semi} &\multirow{7}{*}{CNN-13 (3M)} & - & -                & 11.91$\pm$ 0.22 & 8.83            & 4.21            & - \\
Deep Co-Training~\cite{qiao2018deep}                &             & - & -                &  8.35 $\pm$ 0.06 & -               & 3.29 $\pm$ 0.03 & - \\
Weight Averaging~\cite{athiwaratkun2018improving}   &             & - & 15.58 $\pm$ 0.12 &  9.05 $\pm$ 0.21 & -               & -               & - \\
ICT~\cite{verma2019interpolation}                   &             & - & 15.48 $\pm$ 0.78 &  7.29 $\pm$ 0.02 & 4.78 $\pm$ 0.68 & 3.89 $\pm$ 0.04 & - \\
Label Propagation~\cite{iscen2019label}             &             & - & 16.93 $\pm$ 0.70 & 10.61 $\pm$ 0.28 & -               & -               & - \\
SNTG~\cite{luo2018smooth}                           &             & - & 18.41 $\pm$ 0.52 &  9.89 $\pm$ 0.34 & 4.29 $\pm$ 0.23 & 3.86 $\pm$ 0.27 & - \\
PLCB~\cite{arazo2019pseudo}                         &             & - &  6.85 $\pm$ 0.15 &  5.97 $\pm$ 0.15 & -               & -               & - \\
\midrule
$\Pi$-model~\cite{sajjadi2016regularization}&\multirow{8}{*}{WRN (1.5M)} & 53.02 $\pm$ 2.05 & 31.53 $\pm$ 0.98 & 17.41 $\pm$ 0.37 & 17.65 $\pm$ 0.27 &  8.60 $\pm$ 0.18 & 5.57 $\pm$ 0.14 \\
PseudoLabel~\cite{lee2013pseudo}               & & 49.98 $\pm$ 1.17 & 30.91 $\pm$ 1.73 & 16.21 $\pm$ 0.11 & 21.16 $\pm$ 0.88 & 10.19 $\pm$ 0.41 & 5.71 $\pm$ 0.07 \\
Mixup~\cite{zhang2018mixup}                    & & 47.43 $\pm$ 0.92 & 25.72 $\pm$ 0.66 & 13.15 $\pm$ 0.20 & 39.97 $\pm$ 1.89 & 16.79 $\pm$ 0.63 & 7.96 $\pm$ 0.14 \\
VAT~\cite{miyato2018virtual}                   & & 36.03 $\pm$ 2.82 & 18.68 $\pm$ 0.40 & 11.05 $\pm$ 0.31 &  8.41 $\pm$ 1.01 &  5.98 $\pm$ 0.21 & 4.20 $\pm$ 0.15 \\
Mean Teacher~\cite{tarvainen2017mean}          & & 47.32 $\pm$ 4.71 & 17.32 $\pm$ 4.00 & 10.36 $\pm$ 0.25 &  6.45 $\pm$ 2.43 &  3.75 $\pm$ 0.10 & 3.39 $\pm$ 0.11 \\
MixMatch~\cite{berthelot2019mixmatch}          & & 11.08 $\pm$ 0.87 &  7.75 $\pm$ 0.32 &  6.24 $\pm$ 0.06 &  3.78 $\pm$ 0.26 &  3.27 $\pm$ 0.31 & 2.89 $\pm$ 0.06 \\
ReMixMatch~\cite{berthelot2019remixmatch}      & &  \textbf{6.27 $\pm$ 0.34} &  \textbf{5.73 $\pm$ 0.16} &  5.14 $\pm$ 0.04 &  \textbf{3.10 $\pm$ 0.50} &  \textbf{2.83 $\pm$ 0.30} & \textbf{2.42 $\pm$ 0.09} \\
FeatMatch (Ours)                               & &  7.50 $\pm$ 0.64 &  \textbf{5.76 $\pm$ 0.07} &  \textbf{4.91 $\pm$ 0.18} &  \textbf{3.34 $\pm$ 0.19} &  \textbf{3.10 $\pm$ 0.06} & 2.62 $\pm$ 0.08 \\
\bottomrule
\end{tabular}
}
\end{table*}

%% file: tables/ablation.tex
\begin{table*}[t]
\centering
\caption{
Ablation study on CIFAR-10 with various amount of labeled samples.
}
\label{table:ablation}
\resizebox{0.95\textwidth}{!}{
\begin{tabular}{@{\extracolsep{4pt}}lcccccccc@{}}
\toprule
& & & & & & \multicolumn{3}{c}{\#Labeled samples} \\
\cline{7-9}
Experiment & \makecell{Image-Based\\Augmentation} & \makecell{Feature-Based\\Augmentation} & $\mathcal{L}_{con\text{-}f}$ & $\mathcal{L}_{con\text{-}g}$ & $\mathcal{L}_{con}$ & 250 & 1,000 & 4,000 \\
\midrule
Baseline                           & \CheckmarkBold    & -                 & -                 & -                  & \CheckmarkBold & 19.55 $\pm$ 1.58 & 9.04 $\pm$ 1.00 & 6.08 $\pm$ 0.16 \\
w/o $\mathcal{L}_{con\text{-}f}$   & \CheckmarkBold    & \CheckmarkBold    & -                 & \CheckmarkBold     & -              & 18.57 $\pm$ 3.19 & 8.38 $\pm$ 0.35 & 6.09 $\pm$ 0.16 \\ 
w/o $\mathcal{L}_{con\text{-}g}$   & \CheckmarkBold    & \CheckmarkBold    & \CheckmarkBold    & -                  & -              &  8.19 $\pm$ 1.74 & 6.07 $\pm$ 0.46 & 5.16 $\pm$ 0.30 \\
FeatMatch (Ours)                   & \CheckmarkBold    & \CheckmarkBold    & \CheckmarkBold    & \CheckmarkBold     & -              &  7.90 $\pm$ 0.49 & 5.94 $\pm$ 0.16 & 5.00 $\pm$ 0.21 \\
\bottomrule
\end{tabular}
}
\end{table*}

%% file: sections/conclusion.tex
\section{Conclusion}
We introduce a method to jointly learn a classifier and feature-based refinement and augmentations which can be used within existing consistency-based SSL methods. Unlike traditional image-based transformations, our method can learn complex, feature-based transformations as well as incorporate information from class-specific prototypical representations extracted in an efficient manner (specifically using a memory bank). Using this method, we show comparable results as the current state of the art for smaller datasets such as CIFAR-10 and SVHN, and significant improvements on datasets with a large number of categories (\textit{e.g.}, 17.44\% absolute improvement on mini-ImageNet). We also demonstrate increased robustness to out-of-domain unlabeled data, which is an important real-world problem, and perform ablations and analysis to demonstrate the learned feature transformation and extracted prototypical representations.

\section{Acknowledgement}
This work was funded by DARPA's Learning with Less Labels (LwLL) program under agreement HR0011-18-S-0044 and DARPA’s Lifelong Learning Machines (L2M) program under Cooperative Agreement HR0011-18-2-0019.

%% file: sections/appendix.tex
\section*{Appendix}
\appendix

\renewcommand{\thesection}{\Alph{section}}

\section{State-of-the-art Results with other SSL Techniques}

As we build upon a weaker baseline and our method is much simpler, the performance of our method on the CIFAR-10 dataset is slightly worse in some settings.
However, as we claimed in Section 4.2 of the main paper, our proposed feature-based augmentation method is complementary to conventional image-based augmentation methods and can be easily integrated to further improve the performance.
In Table~\ref{table:soa}, we demonstrate that by incorporating (1) distribution alignment that aligns the marginal class distribution as described in \cite{arazo2019pseudo,berthelot2019remixmatch}, and (2) \emph{Cutout}~\cite{devries2017improved}, an image-based augmentation method, our method indeed compares favorably against current state-of-the-art algorithms.
Note that our method is still simpler when compared to state-of-the-art image-based method, \textit{e.g.,} ReMixMatch~\cite{berthelot2019remixmatch}.
For example, the ReMixMatch method  also incorporates self-supervsied loss, temporal ensembling of model weights, and tailored data augmentation method (CTAugment~\cite{berthelot2019remixmatch}), etc.

\begin{table*}[h]
\begin{center}
\renewcommand{\arraystretch}{1.2}
\caption{
Comparison to other state-of-the-art methods after incorporating some other modern SSL techniques (distribution alignment and Cutout).
We show the results on the CIFAR-10 dataset with varying amounts of labeled samples.
Numbers represent error rate across three runs.
}
\label{table:soa}
\resizebox{0.9\textwidth}{!}{
\begin{tabular}{@{\extracolsep{4pt}}lccc@{}}
\toprule
& \multicolumn{3}{c}{\#Labeled samples} \\
\cline{2-4}
Method & 250 & 1,000 & 4,000 \\
\midrule
MixMatch~\cite{berthelot2019mixmatch} & 11.08 $\pm$ 0.87 & 7.75 $\pm$ 0.32 & 6.24 $\pm$ 0.06 \\
ReMixMatch~\cite{berthelot2019remixmatch} & \textbf{6.27 $\pm$ 0.34} & 5.73 $\pm$ 0.16 & 5.14 $\pm$ 0.04 \\
FeatMatch (Ours in the main paper) & 7.50 $\pm$ 0.64 & 5.76 $\pm$ 0.07 & \textbf{4.91 $\pm$ 0.18} \\
FeatMatch (Ours with other SSL techniques) & \textbf{6.00 $\pm$ 0.41} & \textbf{5.21 $\pm$ 0.08} & \textbf{4.64 $\pm$ 0.11} \\
\bottomrule
\end{tabular}
}
\end{center}
\end{table*}

\section{Pseudo-Labeling Accuracy Before and After $AugF$}
In Section 4.4 of the main paper, we analyze other reasons that $AugF$ improves model performance.
We conclude that our proposed $AugF$ module also learns to refine input feature for a better representation by attending to the prototypes.
This feature refinement process by $AugF$ provides the training objectives of $\mathcal{L}_{con\text{-}g}$ (Eq.~5) and $\mathcal{L}_{con\text{-}f}$ (Eq.~6) with better pseudo-labels, which may be one of the reasons why our method can improve over image-based baseline by a larger margin.
In Fig.~\ref{fig:analysis-pseudolabel} below, we can see that the accuracy of pseudo-labels from the features refined by $AugF$ is higher than those without refinement by $AugF$.

\begin{figure}[h]
\centering
\includegraphics[width=0.8\linewidth]{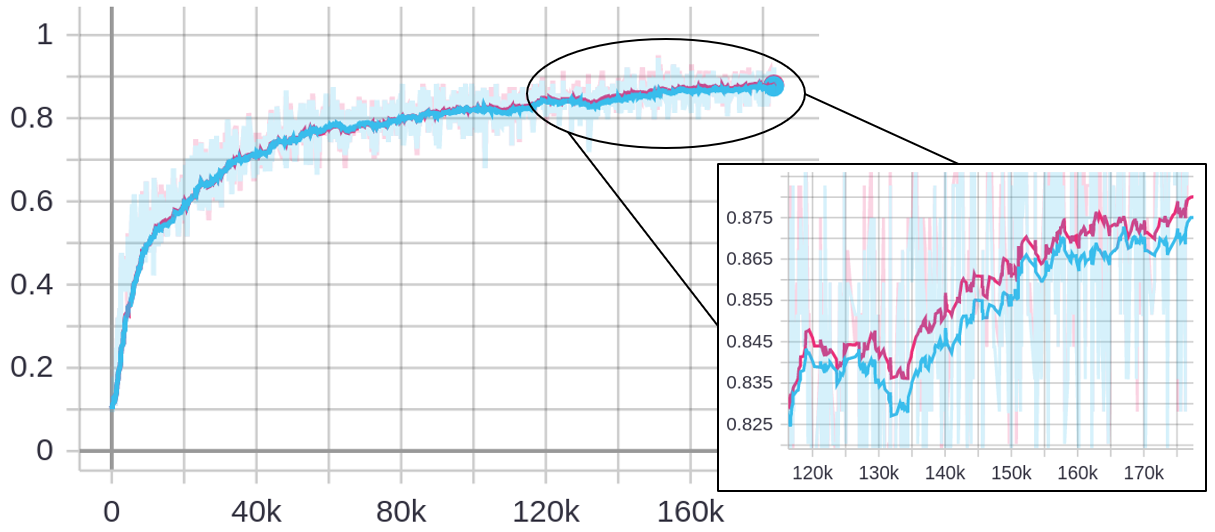}
\caption{
We monitor the accuracy of pseudo-labeling with feature-base refinement (red curve) and without feature-based refinement (blue curve) during training.
We found that the pseudo-label from the refined feature (red) has on average $0.5-1.0 \%$ higher accuracy.
}
\label{fig:analysis-pseudolabel}
\end{figure}

\section{More Analysis on Prototypes}

We test the sensitivity of our method for the hyper-parameter $p_k$ (number of prototypes per class) and $I_p$ (the interval at which a new set of prototypes is extracted).
The analysis is conducted on a held-out validation set of the CIFAR-10 dataset with 250 labels.
As shown in Table~\ref{table:pk}, the final results are stable across different values of $p_k$.
We choose the number of prototypes per class $p_k = 20$ in our method as it performs slightly better than others and has a slightly lower variance.
In Table~\ref{table:ip}, we can see that the final results are also stable across different $I_p$.
Therefore, for simplicity, we extract prototypes every epoch, which is approximately the same as $I_p=400$.

\begin{table*}[h]
\centering
\renewcommand{\arraystretch}{1.2}
\caption{
Sensitivity analysis for $p_k$.
Numbers represent error rates in three runs.
}
\label{table:pk}
\resizebox{0.55\textwidth}{!}{
\begin{tabular}{@{\extracolsep{4pt}}cccc@{}}
\toprule
$p_k=1$ & $p_k=5$ & $p_k=10$ & $p_k=20$ \\
\midrule
8.14 $\pm$ 0.79 & 8.15 $\pm$ 0.19 & 8.01 $\pm$ 0.90 & 8.09 $\pm$ 0.58 \\
\bottomrule
\end{tabular}
}
\end{table*}

\begin{table*}[h]
\centering
\renewcommand{\arraystretch}{1.2}
\caption{
Sensitivity analysis for $I_p$.
Numbers represent error rates in three runs.
}
\label{table:ip}
\resizebox{0.55\textwidth}{!}{
\begin{tabular}{@{\extracolsep{4pt}}cccc@{}}
\toprule
$I_p=200$ & $I_p=400$ & $I_p=600$ & $I_p=800$ \\
\midrule
8.00 $\pm$ 0.81 & 7.99 $\pm$ 0.74 & 7.99 $\pm$ 0.79 & 8.38 $\pm$ 1.21 \\
\bottomrule
\end{tabular}
}
\end{table*}

\section{More Results on the DomainNet Setting}
In Section~4.1, we propose a practical setting where the unlabeled data may come from other domains. 
We show results with different $r_u$, the ratio of unlabeled data coming from the target Real domain or the shifted domains, in Section 4.2.
In this section, we show additional results of $r_u = 0.25$ and $r_u = 0.75$ on both our method and the image-based baseline in Tab.~\ref{table:domainnet_full}.
The results show a similar trend is similar as the Table~3 in the main paper, where the accuracy goes down as $r_u$ goes up.
Our method consistently improves over image-based semi-supervised baseline.
Our method achieves comparable result even in the severe case of $r_u=75\%$ against the image-based baseline method with clean unlabeled data of $r_u=0\%$

\begin{table*}[h]
\centering
\renewcommand{\arraystretch}{1.2}
\caption{
Comparison between the image-based baseline with our proposed feature-based augmentation method on DomainNet with various $r_u$, the ratio of unlabeled data coming from the shifted domains. For instance, $r_u=25 \%$ means 25\% of the unlabeled data are coming from the shifted domains and 75\% are coming from the domain same as the labeled set.
Numbers are error rates across 3 runs, meaning the lower the better.
}
\label{table:domainnet_full}
\resizebox{\textwidth}{!}{
\begin{tabular}{@{\extracolsep{4pt}}lcccc@{}}
\toprule
Method ($5\%$ labeled samples) & $r_u = 0 \%$ & $r_u = 25 \%$ & $r_u = 50 \%$ & $r_u = 75 \%$ \\
\midrule
(Semi-supervised) Baseline  & 56.63 $\pm$ 0.17 & 62.44 $\pm$ 0.67 \% & 65.82 $\pm$ 0.07 & 70.50 $\pm$ 0.51 \\
FeatMatch (Ours)            & \textbf{40.66 $\pm$ 0.60} & \textbf{46.11 $\pm$ 1.15} & \textbf{54.01 $\pm$ 0.66} & \textbf{58.30 $\pm$ 0.93} \\ \hline \hline
Supervised baseline ($5\%$ labeled samples, lower bound) & \multicolumn{4}{c}{77.25 $\pm$ 0.52} \\
Supervised baseline ($100\%$ labeled samples, upper bound) & \multicolumn{4}{c}{31.91 $\pm$ 0.15} \\
\bottomrule
\end{tabular}
}
\end{table*}

\section{Implementation Details}
\subsection{Training}
We train our model with Stochastic Gradient Descent and Nesterov momentum.
As the $AugF$ module heavily relies on the feature representation to compute attention weights, we pre-train the model without $AugF$ for 4 epochs.

We adapt the super convergence learning rate scheduler~\cite{smith2019super} to reduce the total training iterations.
Specifically, in the pre-training stage, the learning rate starts from 4e-4 and linearly increase to 4e-3 in $I_p$ iterations.
After the pre-training stage, we add the $AugF$ module and ramp up the learning rate linearly from 4e-3 to 4e-2 in $I_c$ iterations, and then ramp down back to 4e-3 in another $I_c$ iterations.
In the meantime, the momentum ramps down from 0.95 to 0.85, and then ramps up back to 0.95.
Finally, in the convergence stage, the learning rate ramps further down from 4e-3 to 4e-6 in $I_e$ iterations.

We follow the guidelines in \cite{smith2019super} to set these parameters without aggressive parameter tuning, and set $I_p=3k$, $I_c=75k$, and $I_e=30k$.
As the DomainNet setting has more training samples, we increase these values $I_p=4k$, $I_c=100k$, and $I_e=40k$ without tuning.
We only tune the peak learning rate to be 4e-4 on a held-out validation set on CIFAR-10 with 250 labels.

\subsection{Hyper-parameters}

All the hyper-parameters are tuned on a \emph{held-out validation set} on CIFAR-10 with 250 labels.
These hyper-parameters are shared across all settings and experiments without further tuning.
Since our method is built upon the image-based baseline, we fix the hyper-parameters or select a reasonable value without tuning from the original papers.

\begin{table*}[h]
\centering
\renewcommand{\arraystretch}{1.2}
\caption{Hyper-parameters and their meanings.}
\label{table:hyperparameters}
\resizebox{0.9\textwidth}{!}{
\begin{tabular}{@{\extracolsep{4pt}}clc@{}}
\toprule
Hyper-parameter & Description & Value \\
\midrule
$p_k$ & Number of prototypes per class & 20 \\
$I_p$ & The interval at which a new set of prototypes are extracted & 1 epoch \\
$a_h$ & Number of attention heads in $AugF$ & 4 \\
$\lambda_g$ & Loss weight for $\mathcal{L}_{con\text{-}g}$ & 0.5 \\
$\lambda_f$ & Loss weight for $\mathcal{L}_{con\text{-}f}$ & 2.0 \\
$b_l$ & Batch size for labeled data & 64 \\
$b_u$ & Batch size for unlabeled data & 128 \\
$wd$ & Weight decay & 2e-4 \\
\bottomrule
\end{tabular}
}
\end{table*}

\subsection{Data Augmentation Operations}

We used the same sets of image transformations used in RandAugment~\cite{cubuk2019randaugment}.
There are two parameters in RandAugment: (1) $N$ -- number of operations applied, and (2) $M$ -- maximal magnitude of the applied augmentation.
We use $N=2$ as in RandAugment, and set $M$ to its max value without tuning.
Note that the magnitude is randomly sampled from $[-M, M]$.